\documentclass[sn-mathphys]{sn-jnl}

\makeatletter
\newcommand{\multiline}[1]{%
  \begin{tabularx}{\dimexpr\linewidth-\ALG@thistlm}[t]{@{}X@{}}
    #1
  \end{tabularx}
}
\makeatother

\usepackage{subfiles}
\usepackage{float}
\usepackage[utf8]{inputenc}
\usepackage{dsfont}
\usepackage{amsmath,amssymb,amsfonts,amsthm, bm}
\usepackage{parskip}
\usepackage{breqn}
\usepackage{subcaption,graphicx}
\usepackage{url}
\usepackage{pgfplots}
\usepackage{adjustbox}
\usepackage{tikz}
\usepackage{placeins}
\usetikzlibrary{positioning}
\usepackage{color,soul}
\usepackage{natbib}
\newcommand{\PR}[1]{{\mathbb{P}\left(#1 \right)}}

\begin{document}
\title[XGBoost Quantile Regression]{Composite Quantile Regression With XGBoost Using the Novel Arctan Pinball Loss}

\author*[1]{\fnm{Laurens} \sur{Sluijterman}}\email{l.sluijterman@math.ru.nl}
\author[2]{\fnm{Frank} \sur{Kreuwel}}\email{frank.kreuwel@alliander.com}
\author[1]{\fnm{Eric} \sur{Cator}}\email{e.cator@science.ru.nl}
\author[2]{\fnm{Tom} \sur{Heskes}}\email{tom.heskes@ru.nl}
\affil[1]{\orgdiv{Department of Mathematics}, \orgname{Radboud University}, \orgaddress{\city{Nijmegen}}}
\affil[2]{\orgdiv{System Operations}, \orgname{Alliander}, \orgaddress{\city{Arnhem}}}
\affil[3]{\orgdiv{Institute for Computing and Information Sciences}, \orgname{Radboud University}, \orgaddress{\city{Nijmegen}}}
\abstract{This paper explores the use of XGBoost for composite quantile regression. XGBoost is a highly popular model renowned for its flexibility, efficiency, and capability to deal with missing data. The optimization uses a second order approximation of the loss function, complicating the use of loss functions with a zero or vanishing second derivative. Quantile regression -- a popular approach to obtain conditional quantiles when point estimates alone are insufficient -- unfortunately uses such a loss function, the pinball loss. Existing workarounds are typically inefficient and can result in severe quantile crossings. In this paper, we present a smooth approximation of the pinball loss, the arctan pinball loss, that is tailored to the needs of XGBoost. Specifically, contrary to other smooth approximations, the arctan pinball loss has a relatively large second derivative, which makes it more suitable to use in the second order approximation. Using this loss function enables the simultaneous prediction of multiple quantiles, which is more efficient and results in far fewer quantile crossings.}


\maketitle
\section{Introduction} 
Extreme Gradient Boosting (XGBoost, \citet{chen2016XGBoost}) is a powerful, open-source software library renowned for its performance in structured or tabular data sets across a wide range of domains, including finance \citep{gumus2017Crude, nobre2019Combining}, healthcare \citep{ogunleye2020XGBoost, ramaneswaran2021Hybrid, li2019Gene}, and cybersecurity \citep{dhaliwal2018Effective, jiang2020Network}. XGBoost is increasingly being used for safety-critical applications, such as predicting floods \citep{ma2021XGBoostbased}.

For these safety-critical applications, it is typically insufficient to rely solely on predicting a point estimate. When predicting the water level in a river, understanding the potential extreme values is more important than predicting averages. Quantile regression \citep{koenker1978regression} offers an attractive solution. Instead of merely predicting a point estimate, various quantiles are predicted, for instance including the 0.95-quantile of the water levels. 

Quantile regression has been carried out with a large variety of models. While originally mainly linear models were used, modern implementations of quantile regression often leverage complex models such as random forest \citep{meinshausen2006Quantile} and neural networks \citep{hatalis2019Novela}. These implementations may also predict multiple quantiles with a single model \citep{xu2017composite}, typically referred to as composite quantile regression. 

A large advantage of quantile regression is that it makes no distributional assumptions. Many uncertainty estimation methods will typically assume a Gaussian distribution \citep{lakshminarayanan2017simple, nix1994estimating, gal2016dropout}, which could lead to subpar prediction intervals if the data is not normally distributed. Given the large popularity of XGBoost and quantile regression, there is a clear appeal to use XGBoost for this task.

Unfortunately, using XGBoost for quantile regression is nontrivial. At its heart, the model uses a quadratic approximation of the loss function during the optimization. However, as we will discuss in more detail later, the objective that is typically used for quantile regression, the pinball loss, is not differentiable everywhere and has a second derivative of zero, which makes this second-order approximation impossible.

Several solutions have been developed. The current implementation in the XGBoost package uses a different type of trees, additive trees, that do not require the second derivative during the optimization. However, this requires the use of separate models for each quantile. This is undesirable both as this can easily result in a very high number of quantile crossings -- for example, the 0.45-quantile being larger than the 0.55-quantile --  and because it is inefficient \citep{zou2008Composite}. Another option is to use the regular XGBoost model but with a smooth approximation of the pinball loss that is differentiable everywhere.

While various of these approximations have been used for neural networks \citep{hatalis2019Novela, zheng2011Gradient, xu2017composite}, these approximations typically have a second derivative that is either zero or becomes extremely small. These approximations are unsuitable for XGBoost given its reliance on the second-order approximation of the loss function.

In this paper, we therefore present a novel smooth approximation, named the arctan pinball loss, specifically tailored for XGBoost. Crucially, the loss function is differentiable everywhere and has a much larger second derivative than the existing alternatives, making it more suitable for XGBoost. This allows the use of a single model for multiple quantiles, resulting in far fewer crossings and an increased efficiency. 

Our paper is organized as follows. Section \ref{backgroundrelatedwork} contains all relevant technical details on XGBoost and quantile regression. Additionally, the existing smooth approximations are discussed. The arctan pinball loss is presented in Section \ref{arctanlosssection}. In Section \ref{results}, our implementation of quantile regression with XGBoost is compared to the current implementation. Crucially, our approach has significantly fewer crossings while achieving similar or superior coverage. Final concluding remarks can be found in Section \ref{conclusionsection}.

\section{Background and related work} \label{backgroundrelatedwork}
This section consists of three parts. We first provide the details on XGBoost that are necessary for this paper. We then discuss quantile regression and explain why it is non-trivial to use XGBoost for this task. The third subsection provides current solutions for this problem along with the shortcomings of those solutions.
\subsection{XGBoost}
We provide an introduction to XGBoost at the minimal level that is required for this paper. For a more in-depth introduction, we refer to \citet{chen2016XGBoost}, whose notation we have followed here. 

XGBoost is a boosting approach \citep{schapire1990Strength} that iteratively trains weak learners, typically tree models, while employing both the first and second derivative of the loss function, hence the name Extreme Gradient.

The eventual output of the model is the sum of the outputs of the $K$ trees:
\begin{equation}
\hat{y}_{i} = \phi(\bm{x}_{i}) = f_{0}(\bm{x}_{i}) + \eta \sum_{k=1}^{K} f_{k}(\bm{x}_{i}),	
\end{equation}
where $f_{0}(\bm{x}_{i})$ is the base score, $\eta$ is the learning rate, and $f_{k}$ is a tree with tree structure $q_{k}$, a function that maps the inputs to a leaf index, and weights-vector $\bm{\omega}_{k}$, a vector containing the weights of each leaf.  

XGBoost iteratively trains the trees with the goal to predict the remaining residual. These individual trees are trained by optimizing a regularized objective:
\begin{equation}
\mathcal{L}(\phi) = \sum_{i} l(y_{i}, \hat{y}_{i}) + \sum_{k}\Omega(f_{k}), \quad \text{with} \; \; \Omega(f_{k}) = \gamma T_{k} + \frac{1}{2} \lambda ||\bm{\omega}_{k}||^{2}.
\end{equation}
The loss function, $l$, measures the difference between the outputs, $\hat{y}$, and the observations, $y$. The output could be an estimate of $y$ but it could also be a conditional quantile. The regularisation term $\Omega$ favors simpler trees with a smaller number of leaves, $T_{k}$, and smaller weights. 

The model is trained iteratively, one tree at a time. Each tree aims to learn the residual from all the previous trees. Let $\hat{y}_{i}^{(t)}$ be the $i$-th prediction after having trained the first $t$ trees. During the training of tree $t$, the following objective is optimized:
\begin{equation}
\mathcal{L}^{(t)} = \sum_{i} l(y_{i}, \hat{y}_{i}^{(t-1)} + f_{t}(\bm{x}_{i})) + \Omega(f_{t}).
\label{eq: Lt}
\end{equation}
XGBoost uses a quadratic approximation of Equation \eqref{eq: Lt} during the optimization:
\begin{equation}
\mathcal{L}^{(t)} \approx \sum_{i}[ l(y_{i}, \hat{y}_{i}^{(t-1)}) + f_{t}(\bm{x}_{i})g_{i}  + \frac{1}{2} f_{t}^{2}(\bm{x}_{i})h_{i}] + \Omega(f_{t}),
\label{eq: approxLt}
\end{equation}
where $g_{i} = \frac{\partial l(y_{i}, \hat{y}^{(t-1)}_{i})}{\partial \hat{y}_{i}^{(t-1)}}$, and $h_{i} =  \frac{\partial^{2} l(y_{i}, \hat{y}^{(t-1)}_{i})}{\partial (\hat{y}_{i}^{(t-1)})^{2}}$.
Using this equation, the optimal weight of leaf $j$ of tree $t$ can be calculated:
\begin{equation}
\omega_{tj}^{\ast} = - \frac{\sum_{i \in I_{tj}} g_{i}}{\sum_{i \in I_{tj}} h_{i} + \lambda},	
\label{eq: optimalweight}
\end{equation}
where $I_{tj} = \{i|q_{t}(\bm{x}_{i}) = j \}$, the indices of the data points that end up in leaf $j$.

By using Equation \eqref{eq: optimalweight}, the approximate loss function in Equation \eqref{eq: approxLt} can be calculated for a specific tree structure. An efficient split-finding algorithm is used to find the optimal tree structure.

XGBoost further distinguishes itself through several key features that enhance its performance and versatility in machine learning tasks. Firstly, it employs a highly efficient split finding algorithm that optimizes the selection of split points in trees, significantly speeding up the learning process. Secondly, XGBoost has excellent parallelization capabilities, allowing it to utilize multiple cores during the training phase, which greatly reduces the time required to build models. Furthermore, it is adept at handling missing values in the data set. XGBoost automatically learns the best direction to assign missing values during the split, either to the left or right child, depending on which choice leads to the best gain. This ability to deal with incomplete data directly, without needing imputation or dropping rows, makes XGBoost a robust and flexible tool for a wide array of data science and machine learning applications.

\subsection{Quantile regression}
\begin{figure}[h!]
\centering
\includegraphics[width=0.5 \textwidth]{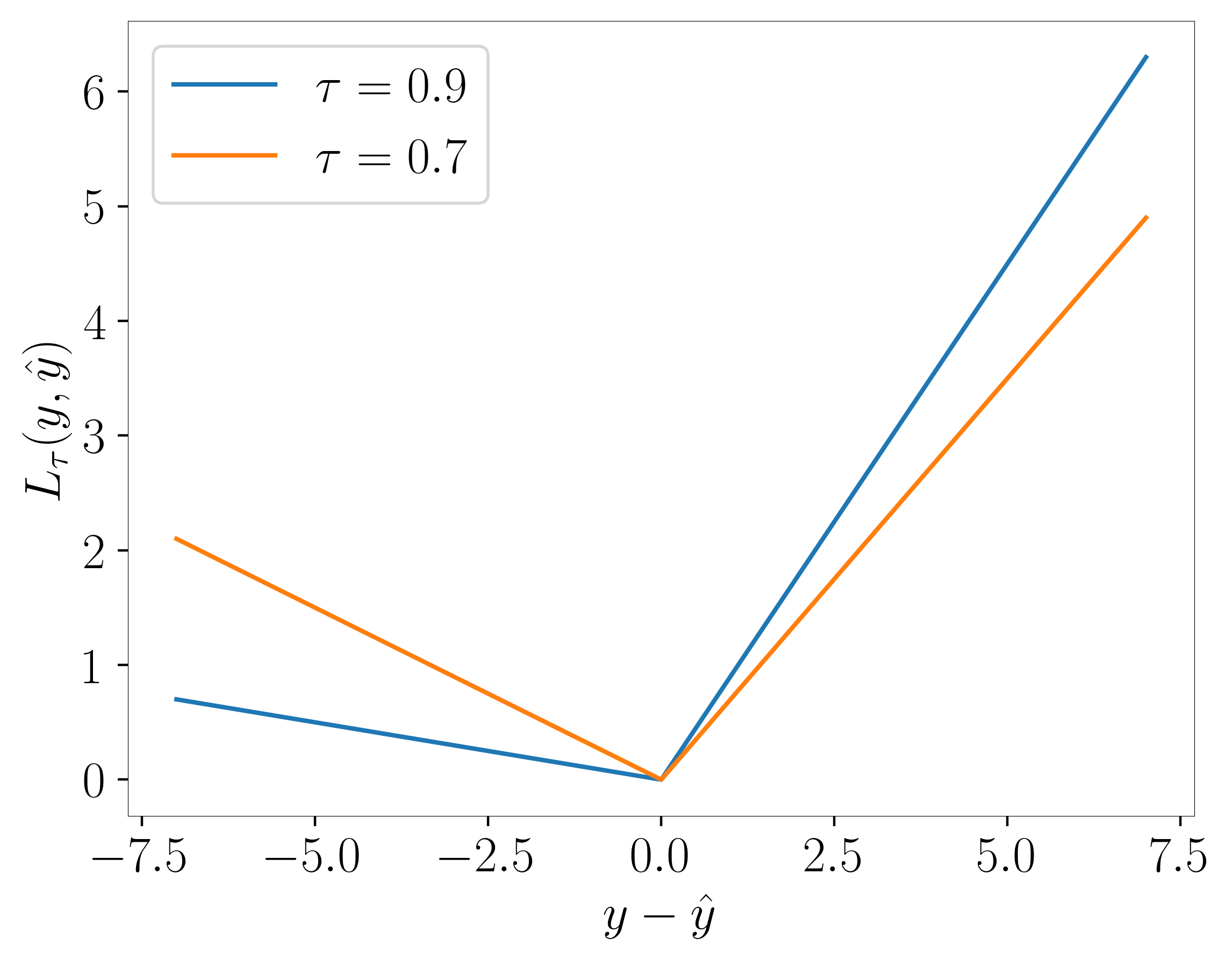}	
\caption{The pinball loss for two different values of $\tau$.}
\label{fig: pinballloss}
\end{figure}

Quantile regression aims to predict a specific quantile of a probability distribution rather than, for instance, predicting the mean. A conditional quantile is defined as:
\[q_{\tau}(\bm{x}) := \min \{y | F_{Y \mid X=\bm{x}}(y)\geq \tau \}.
\]
In other words, it is the smallest value $y$, such that the probability that $Y$ is smaller than $y$, given $X=\bm{x}$, is at least $\tau$. \citet{koenker1978regression} showed that conditional quantiles can be estimated by minimizing the  pinball loss: 
\begin{equation}
L_{\tau}(y_{i}, \hat{y}_{i}) = \tau(y_{i} - \hat{y}_{i}) \mathbb{I}_{\{\hat{y}_{i} \leq y_{i}\}} +(\tau -1)(y_{i} - \hat{y}_{i})\mathbb{I}_{\{\hat{y}_{i} > y_{i}\}},
	\label{eq: pinball loss}
\end{equation}
where $\hat{y}_{i}$ is the predicted quantile and $y_{i}$ is the observed value. The pinball loss is visualized in Figure \ref{fig: pinballloss} for two different values of $\tau$. The intuition behind the loss is that for the 0.9-quantile, estimating a quantile smaller than the observation is penalized more than estimating a quantile that is too large.

Note that the pinball loss only depends on $y_{i} - \hat{y}_{i}$. In the remainder of this paper, we will therefore use the notation $u := y - \hat{y}$ and present the loss functions in terms of $u$.

Ideally, we would want to let XGBoost output the quantiles and use the pinball loss function directly. However, this is not possible for two reasons. First of all, the loss function is not differentiable at $u=0$. Secondly, the second derivative is zero everywhere. This is problematic for XGBoost since it uses a second order approximation of the loss function during the optimization.

\subsection{Previous solutions}
The current solution in the XGBoost package is to use additive trees. These are slightly different trees that do not require the second derivative but rely on an adapted training algorithm that uses line searches. However, using these modified trees requires a separate model for each quantile, which is highly inefficient.

The problem of the differentiability at $u=0$ can also be overcome by using a smooth approximation of the loss function. Multiple different smooth approximations have been suggested for neural networks. 

\citet{hatalis2019Novela} use the following smooth approximation based on work from \citet{zheng2011Gradient}:
\begin{equation}
	L^{(\text{exp})}_{\tau, s}(u) = \tau u + s \log\left(1 + \exp(- u / s)\right),
\label{eq: exppinballloss}
\end{equation}
where $s$ is a smoothing parameter that determines the amount of smoothing. A smaller value gives a closer approximation to the true pinball loss.

\citet{cannon2011quantile} and \citet{xu2017composite} use the Huber norm to approximate the pinball loss. The Huber norm is given by:
\begin{equation}
n_{\delta}(u) = 
\begin{cases} 
\frac{1}{2} u^2 & \text{for } |u| \leq \delta, \\
|u| - \frac{1}{2} \delta & \text{otherwise}.
\end{cases}	
\end{equation}
The resulting approximation of the pinball loss is given by:
\begin{equation}
	L^{(\text{Huber})}_{\tau, \delta}(u) = \tau n_{\delta}(u) \mathbb{I}_{u > 0} +(1-\tau )n_{\delta}(u) \mathbb{I}_{u<0}.
	\label{eq: pinball loss}
\end{equation}

This Huber pinball loss has also been applied to XGBoost \citep{yin2023Quantile}. However, since the second derivative of the Huber pinball loss is still zero for $|u| > \delta$, the algorithm requires a large value of $\lambda$ to properly converge in practice. Being forced to use a large $\lambda$ is undesirable. The second derivative becomes obsolete and the training in practice reduces to gradient descent with a very low learning rate. This can be seen by evaluating Equation \eqref{eq: optimalweight} for a large value of $\lambda$. 

To really benefit from the higher convergence speed achieved by the quadratic approximation used by XGBoost, it is essential to use an approximation of the pinball loss that is not only differentiable at zero but also has a non-zero second derivative everywhere. The exponential approximation, $L^{(\text{exp})}_{\tau, s}(u)$, may therefore seem like a suitable candidate. In fact, the second derivative is strictly positive:

\begin{equation}
\frac{\partial^{2}L^{(\text{exp})}_{\tau, s}(u)}{\partial u^{2}} = 	\left(\exp(-\frac{u}{2s}) + \exp(\frac{u}{2s})\right)^{-2}  \frac{1}{s}.
\end{equation}
However, when implementing this, we ran into similar problems. Although the second derivative is always positive, it decays exponentially as a function of $|u|$, resulting in a vanishing second derivative. In summary, we need to find a smooth approximation with a reasonably large second derivative.

\section{The arctan pinball loss} \label{arctanlosssection}
Our goal is to develop a smooth approximation of the pinball loss function that maintains a large second derivative. To achieve this, we introduce the following approximation, named the arctan pinball loss:
\begin{equation}
L^{(\text{arctan})}_{\tau, s}(u) = \left(\tau - 0.5 + \frac{\arctan (u/s)}{\pi}\right)u	 + \frac{s}{\pi},
\end{equation}
where $s$ is a smoothing parameter that controls the amount of smoothing. A smaller value of $s$ results in a closer approximation but, as we will soon see, also a smaller second derivative. The $s/\pi$ term ensures that the approximation is unbiased for large values of $|u|$. For XGBoost, this term is purely aesthetic, as it does not influence the first or second derivative. However, for other applications or optimisation procedures, it could be useful. We provide more details on the construction and the unbiasedness in Appendix \ref{arctanconstruction}.

 The second derivative of the arctan pinball loss is given by:
\begin{align*}
\frac{\partial^{2}L^{(\text{arctan})}_{\tau, s}(u)}{\partial u^{2}} &= 	\frac{2}{\pi s}\left(1 + (u/s)^{2}\right)^{-1} - \frac{2u^{2}}{\pi  s^3}  \left(1 + (u/s)^{2}\right)^{-2} \\
&= \frac{2}{\pi s}(1 + (u/s)^{2})^{-2}.
\end{align*}
Crucially, this second derivative is strictly positive and falls off polynomially as opposed to the exponential decay of $\frac{\partial^{2}L^{(\text{exp})}_{\tau, s}(u)}{\partial u^{2}}$.

\begin{figure}[tb]
    \centering
    \begin{subfigure}[b]{0.45\textwidth} 
        \centering
        \includegraphics[width=\textwidth]{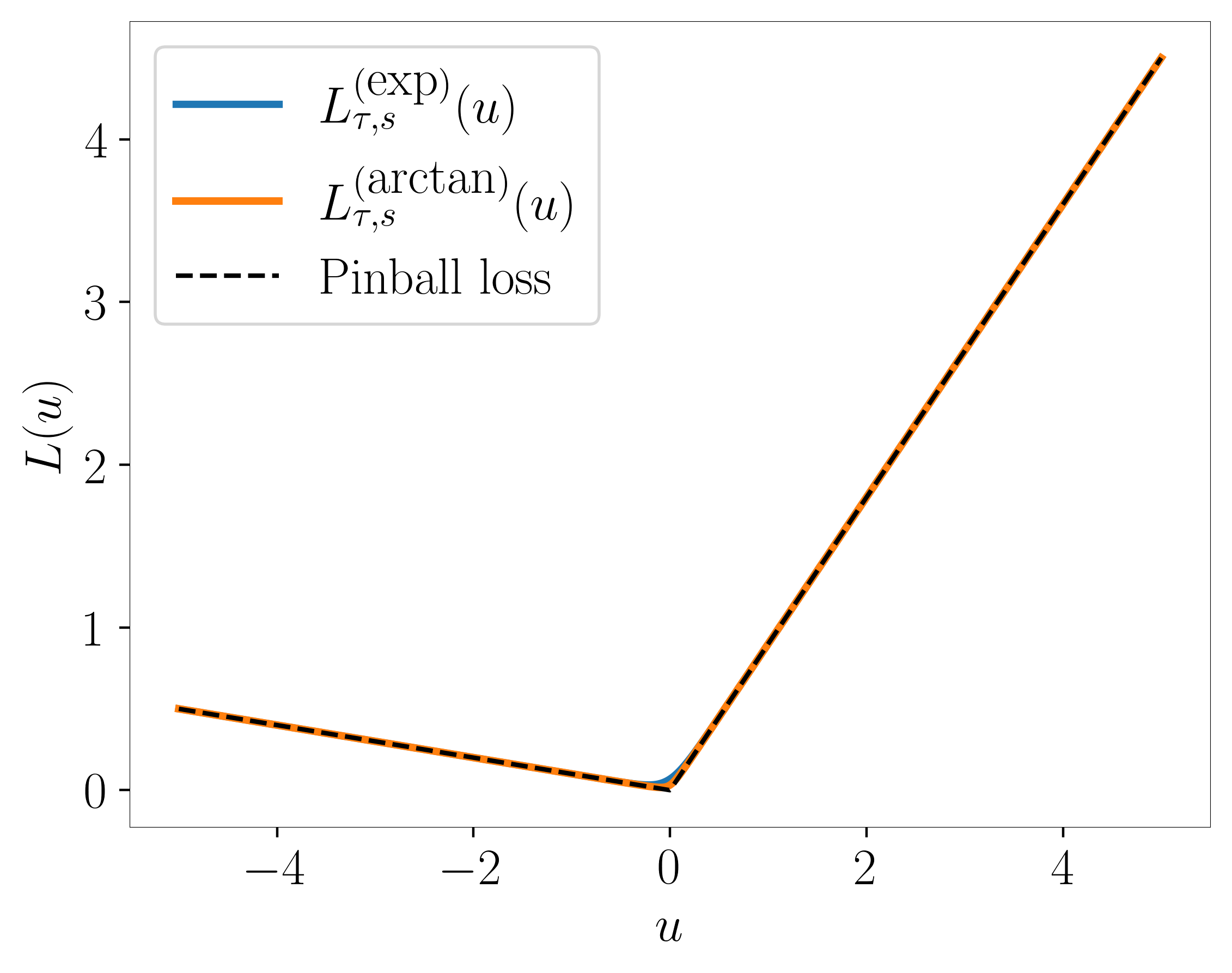} 
        \caption{}
        \label{fig:sub1}
    \end{subfigure}
    \hfill 
    \begin{subfigure}[b]{0.45\textwidth}
            \centering
        \includegraphics[width=\textwidth]{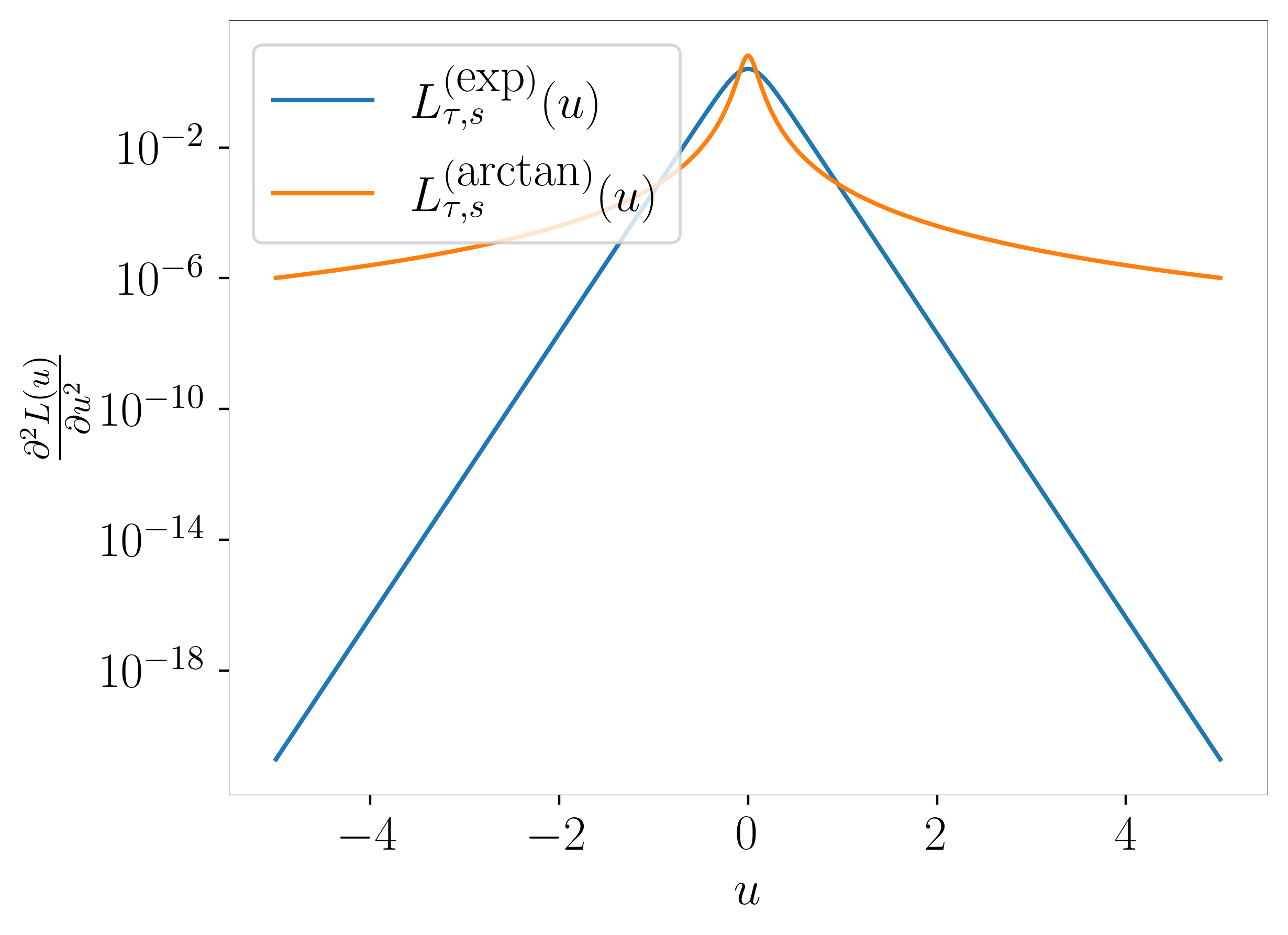} 
          \caption{}
        \label{fig:sub2}
    \end{subfigure}
    \caption{A comparison of $L^{(\text{exp})}_{\tau, s}(u)$ and $L^{(\text{arctan})}_{\tau, s}(u)$ for $\tau=0.9$ and $s=0.1$. Both the exponential approximation and the arctan approximation approximate the pinball loss very closely. However, as is displayed in (b), the second derivative of the arctan pinball loss is much larger.}
    \label{fig: losscomparison}
\end{figure}

Figure \ref{fig: losscomparison} visualizes this difference in second derivative. Figure \ref{fig: losscomparison}(a) shows that both the exponential pinball loss and the arctan pinball loss approximate the true pinball loss very well when using $s=0.1$. However, as can be seen in Figure \ref{fig: losscomparison}(b), the arctan pinball loss has a second derivative that is orders of magnitude larger, making it a much better candidate to use with XGBoost.

By using this loss function, we are able to carry out quantile regression while using the default version of XGBoost. One of the advantages of this is that we can predict multiple quantiles with the same model by using multi-output leaves. From a theoretical point of view, using the same model for multiple quantiles is advantageous. The different quantiles can share information, making it more efficient than estimating all the quantiles with separate models \citep{zou2008Composite}.

A second advantage of using the same model for different quantiles is that all these quantiles share the same splits. This makes it much less likely that quantiles cross. As we will see in the Section \ref{results}, using separate models for each quantile results in many more quantiles crossings, which is clearly undesirable.

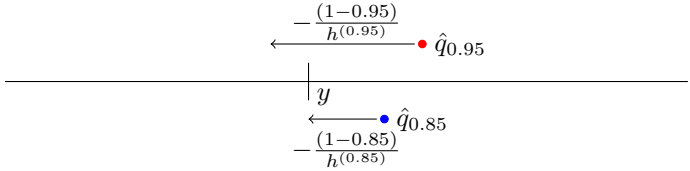
\begin{figure}[tb]
\begin{center}
\begin{tikzpicture}
\draw node at (4.2, 2) {\textbf{Situation 1}: \text{Different second derivative and gradient}};
	\draw (1,0) -- (10, 0.0);
	\draw (5, -0.25) -- (5, 0.25);
	\draw[->] (1.1, -0.5) -- (6, -0.5);
	\draw node at (3.5, -0.8) {$ \frac{0.85}{h^{(0.85)}}$};
	
	\draw node at (5.2, -0.2) {$y$};
	\draw node at (0.5, -0.5) {$\hat{q}_{0.85}$};
	\draw node at (7.5, 0.5) {$\hat{q}_{0.95}$};
	
	\filldraw[blue] (1, -0.5) circle (0.05);
	\filldraw[red] (7, 0.5) circle (0.05);
	\draw node at (6.3, 0.8) {$ -\frac{(1-0.95)}{h^{(0.95)}}$};
	\draw[->] (6.9, 0.5) -- (5.5, 0.5);	
\draw node at (2.5, -2) {\textbf{Situation 2}: \text{Different gradient}};
	\draw (1,-4) -- (10, -4);
	\draw (5, -4.25) -- (5, -3.75);

	\draw node at (5.2, -4.2) {$y$};

	\filldraw[blue] (3.5, -4.5) circle (0.05);
	\draw[->] (3.6, -4.5) -- (6.0, -4.5);
	\draw node at (4.67, -4.8) {$\frac{0.85}{h^{(0.85)}}$};
	\draw node at (3.0, -4.5) {$\hat{q}_{0.85}$};
	
	\filldraw[red] (6.5, -3.5) circle (0.05);
	\draw[->] (6.4, -3.5) -- (5.5, -3.5);
	\draw node at (7, -3.5) {$\hat{q}_{0.95}$};	
	\draw node at (5.9, -3.2) {$ -\frac{(1-0.95)}{h^{(0.95)}}$};

\draw node at (3.2, -6) {\textbf{Situation 3}: \text{Different second derivative}};
	\draw (1,-8) -- (10, -8);
	\draw (5, -8.25) -- (5, -7.75);
	\draw node at (5.2, -8.2) {$y$};

	\filldraw[blue] (6.0, -8.5) circle (0.05);
	\draw[->] (5.9, -8.5) -- (5.0, -8.5);
	\draw node at (5.5, -8.9) {$-\frac{(1-0.85)}{h^{(0.85)}}$};
	\draw node at (6.5, -8.5) {$\hat{q}_{0.85}$};
	
	\filldraw[red] (6.5, -7.5) circle (0.05);
	\draw[->] (6.4, -7.5) -- (4.5, -7.5);
	\draw node at (7, -7.5) {$\hat{q}_{0.95}$};	
	\draw node at (5.5, -7.2) {$ -\frac{(1-0.95)}{h^{(0.95)}}$};
\end{tikzpicture}
\end{center}
\caption{Three scenarios where crossings can occur, not at scale. While the optimum of the arctan pinball loss has no crossings, individual updates can result in crossings due to the quadratic approximation. The resulting update is proportional to the gradient divided by the second derivative, denoted with $h$.}
\label{fig: crossings}
\end{figure}

However, even when using a single model, crossings cannot be entirely prevented. Due to the quadratic approximation, a single update can still result in a crossing. Three scenarios where crossings could occur during an update are visualized in Figure \ref{fig: crossings}.

For simplicity, we consider the scenario where there is only a single data point in a leaf. Suppose we predict the 0.95-quantile (red) and the 0.85-quantile (blue). Without any regularization, $\lambda=0$, the update for both quantiles is proportional to the gradient divided by the second derivative (Equation \eqref{eq: optimalweight}). 

In situation 1, the 0.95-quantile is slightly larger than $y$ and the 0.85-quantile substantially smaller than $y$. The update is proportional to the gradient divided by the second derivative. In the first situation, the gradient for the 0.95-quantile is smaller than for the 0.85-quantile and the second derivative is larger. These resulting updates cause the 0.85-quantile to become substantially bigger and the 0.95-quantile to become slightly smaller. This could result in a crossing. 

In situation 2, the 0.85-quantile is smaller than $y$ and the 0.95-quantile is larger than $y$ by a similar amount. In this case, the second derivatives for both are equal. However, the gradient for the 0.85 quantile is roughly 0.85 compared to -0.05 for the 0.95 quantile. This could also result in a crossing during this update. 

In the final scenario, both quantiles are larger than $y$. The gradient of the 0.95-quantile is $-0.05$ and the gradient for the 0.85-quantile is -0.15. At first glance this should not be able to result in a crossing. However, since the second derivative for $\hat{q}_{0.95}$ is smaller, this is still a possibility.

Note that two of the three crossing scenarios were caused by a difference in second derivative. Since the second derivative of our arctan pinball loss is polynomial instead of exponential, we do not suffer from this effect as much. Additionally, using a larger $\lambda$ would also diminish this effect.

\begin{figure}[tb]
\centering
    \begin{subfigure}[b]{0.45\textwidth} 
        \centering
        \includegraphics[width=\textwidth]{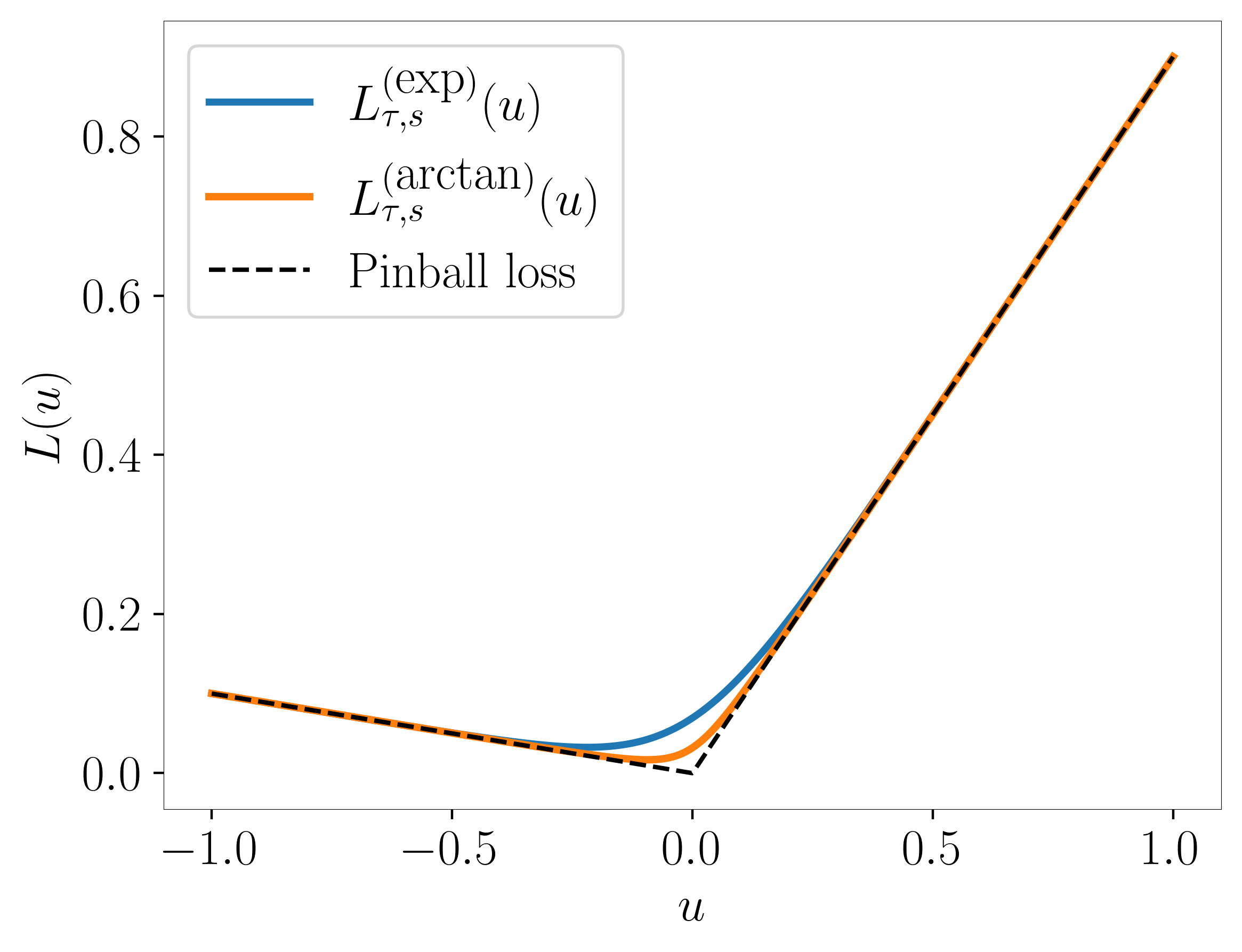} 
        \caption{}
        \label{fig:sub1}
    \end{subfigure}
    \hfill 
    \begin{subfigure}[b]{0.45\textwidth}
            \centering
        \includegraphics[width=\textwidth]{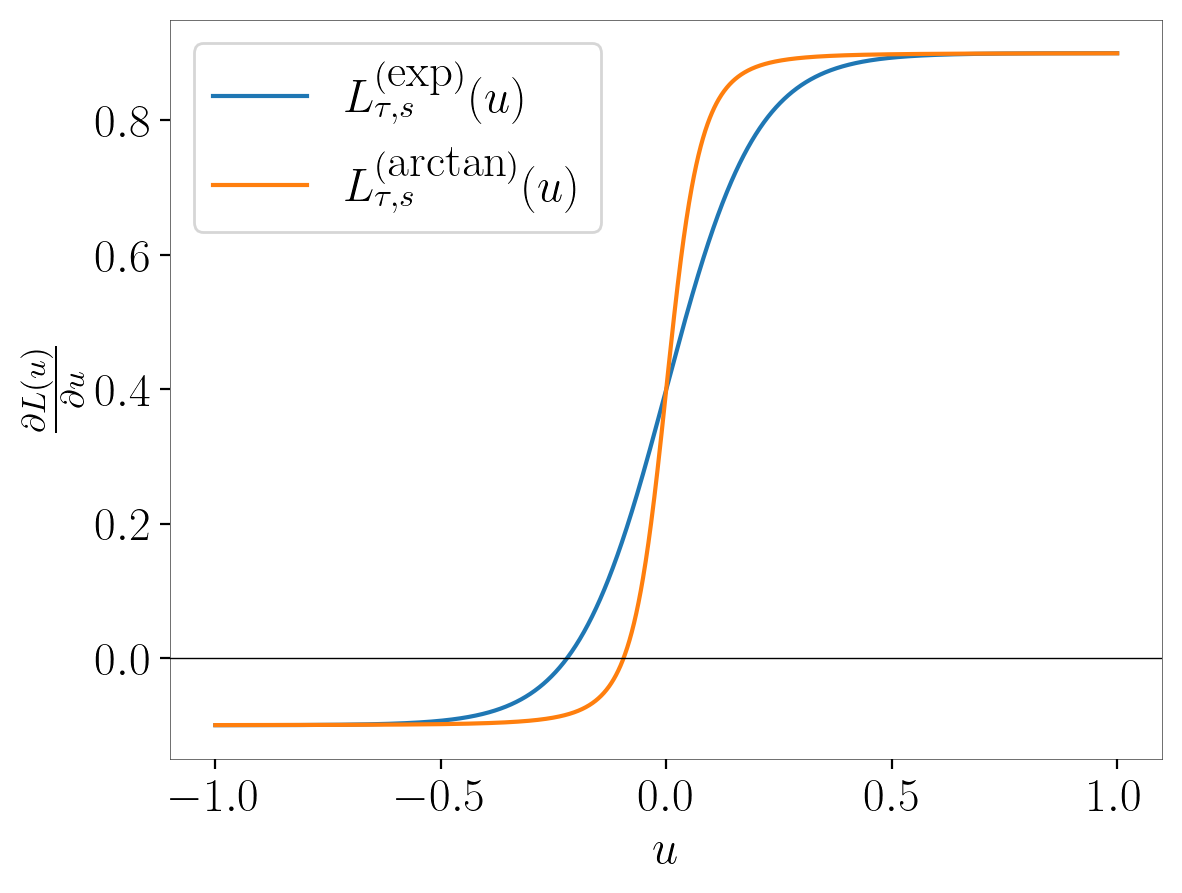} 
          \caption{}
        \label{fig:sub2}
    \end{subfigure}
\caption{A comparison of $L^{(\text{exp})}_{\tau, s}(u)$ and $L^{(\text{arctan})}_{\tau, s}(u)$ for $\tau=0.9$ and $s=0.1$ at a very small scale. Close to the origin, the bias in both approximations is clear. Both approximations have the actual minimum of the loss function slightly below $u=0$. This results in slightly more conservative quantiles, meaning larger quantiles for $\tau > 0.5$ and smaller quantiles for $\tau<0.5$ compared to when using the regular pinball loss. This effect is larger when using a larger value of $s$.}
\label{fig: losscomparisonzoom}
\end{figure}

In general, using any approximation of the true loss can result in a slightly biased model. Figure \ref{fig: losscomparisonzoom} illustrates the bias that both approximations of the pinball loss, $L^{(\text{exp})}_{\tau, s}(u)$ and $L^{(\text{arctan})}_{\tau, s}(u)$, have near the origin. The optimum for both losses is slightly below $u=0$ when using a $\tau$ larger than 0.5. This causes the predicted quantiles to be slightly larger. This would result in slightly more conservative prediction intervals, especially when using larger values of $s$. We will observe this behaviour in Section \ref{results}.

For optimal use of the arctan pinball loss, we recommend the following modeling choices.
\begin{enumerate}
	\item Always use standardized targets. This allows us to keep certain hyper-parameters,  most notably the smoothing parameter $s$, fixed regardless of the data set. We typically found values between 0.05 and 0.1 to work well. Smaller values result in extremely small second derivatives, and much larger values result in an approximation that is too rough, leading to overly conservative prediction intervals.
	\item Set the min-child-weight parameter to zero. This parameter regularizes the trees by requiring a minimum weight in each leaf in order to allow a split. The weight is defined as the sum of the second derivatives of the points in the resulting leaf. This makes sense when using a loss function with a constant second derivative, such as the mean-squared error. In that case, this parameter enforces a minimum number of data points in each leaf to prevent overfitting. However, since the second derivative of our loss function is far from constant, we advise to not use this parameter and set it to zero.
	\item Use a slightly smaller learning rate of 0.05 (compared to 0.1 in the standard implementation). The weights of the new tree are given by Equation \eqref{eq: optimalweight}. The outputs of the new tree are multiplied by the learning rate to obtain the actual update. Since the second derivative can be substantially smaller than 1, it is still possible to obtain rather large updates. To make this more stable, we advise to use a slightly smaller learning rate.
	\item Set the max-delta parameter to 0.5. This is done for the same reason as the slightly lower learning rate. To prevent overly large updates, this parameter is set to 0.5. During our experiments, we observed no negative effects of using this parameter in terms of coverage or validation loss but it reduced the number of quantile crossings.
\end{enumerate}
\section{Experimental results} \label{results}
This section consists of three parts. We first go through the various data sets that were used. Subsequently, we explain our experimental design. This includes the choices of hyper-parameters, the optimization procedure, and the metrics that were used. Finally, the results are given and discussed in the third subsection. Our implementation of the arctan pinball loss is publicly available: \url{https://github.com/LaurensSluyterman/XGBoost_quantile_regression}.
\subsection{Data sets}
\paragraph{Toy example} 
Our first example is a one-dimensional toy example. This experiment demonstrates the qualitative advantages of our approach. Specifically, we illustrate that the splits for the different quantiles are all located at the same positions, significantly reducing the number of quantile crossings.

The training set consists of 1,000 realizations of the random variable pair $(X, Y)$, where $X\sim U[0,1]$ and $Y \mid X = x \sim \mathcal{N}(\sin(7x), 0.2^2)$. 

\paragraph{UCI benchmark data sets} 
Secondly, we examine our method on six publicly available UCI regression data sets: Boston housing, energy, concrete, wine quality, yacht, and kin8nm. These data sets range from a few hundred data points, with yacht being the smallest at 308, to several thousands, kin8nm having over 8,000 data points. The data sets feature between 6 and 13 covariates, encompassing both continuous and categorical variables. Given their high dimensionality and the wide range of tasks they represent, these data sets are frequently used as benchmark data sets in machine learning research \citep{hernandez2015probabilistic}.

\paragraph{Electricity-grid substations}
Lastly, we examine the total load on four distinct substations from the Dutch electricity grid. For each substation, three months of data at a temporal resolution of 15 minutes is available. The objective is to predict the load on the substation one day ahead using the 81 available covariates. These covariates comprise a mix of measurements, predictions, and categorical values. Examples include load measurements from the previous day, day-ahead electricity price, predicted amounts of solar radiation and windspeed for the next day, and calendar-derived variables such as whether the day is a weekday or a holiday. These data sets have been provided to us by the distribution system operator Alliander and are publicly available as part of the OpenSTEF package: \url{https://github.com/OpenSTEF/openstef-offline-example/tree/master/examples/data}.

\subsection{Experimental design}
For all experiments, we predict 10 different quantiles: 
\[[0.05, 0.15, 0.25, \ldots, 0.75, 0.85, 0.95].\]

The following hyper-parameters are optimized:
\begin{itemize}
	\item The number of estimators: [100, 200, 400].
	\item The $\lambda$ regularization parameter: [0.01, 0.1, 0.25, 0.5, 1, 2.5, 5, 10].
	\item The $\gamma$ regularization parameter: [0.1, 0.25, 0.5, 1, 2.5, 5, 10].
	\item The maximum depth of the trees: [2, 3, 4].
\end{itemize}

For the toy example, we applied 3-fold cross-validation to determine the optimal hyper-parameters and evaluated the resulting model on a separate test set.

For the UCI data sets, we used 3-fold cross-validation to obtain predicted quantiles for every data point in the data set. During each cross-validation, another round of 3-fold cross-validation was used to determine the optimal hyper-parameters.

\begin{figure}[h!]
\begin{center}
	\begin{tikzpicture}[scale=0.7]
    \fill[blue!20] (0,0) rectangle (12,1);
    \node at (7.5,0.5) {Training};

    \fill[red!20] (12,0) rectangle (13.5,1);
    \node at (12.75,0.5) {Val};

    \fill[green!40] (13.5,0) rectangle (15,1);
    \node at (14.25,0.5) {Test};

    \draw (0,0) rectangle (15,1);
    
    \draw[->, thick] (6, -1) -- (9, -1) node[above, pos=0.5]{$t$};
	\end{tikzpicture}
\end{center}
\caption{Illustration of the train/val/test split used for the Alliander data.}
\label{fig: alliandersplit}
\end{figure}
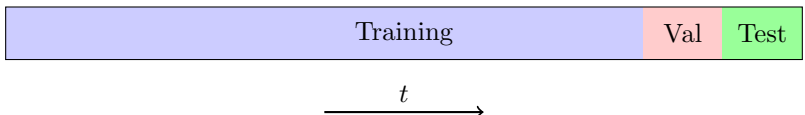

Since the substation data sets are time series, we could not use regular cross-validation. Instead, we used a train/validation/test split where we allocated the first 80$\%$ of the time series as the training set, the next 10$\%$ as the validation set, and the final 10$\%$ as the test set. The optimal hyper-parameters were determined using the validation set, and the actual model was fitted using these parameters on the combined training and validation set. Subsequently, the model was evaluated on the test set. This procedure is visualized in Figure \ref{fig: alliandersplit}.
\paragraph{Metrics} For the toy-example, which is mainly illustrative, we  provide visualizations of the various quantiles. For the UCI data set and the electricity-grid substation data sets, we provide the following quantitative metrics:

1. The marginal coverage percentage and average width of the 90$\%$ PI:
\begin{equation}
\text{Coverage} = \frac{1}{ N} \sum_{i=1}^{N}\mathbb{I}_{\{y_{i} \in \text{PI}(\bm{x}_{i})\}}\cdot 100\%,
\end{equation}
where $\text{PI}(\bm{x}_{i})$ is the 90$\%$ PI that is constructed using the predicted 0.05- and 0.95-quantile. We also report the average width of this interval.

The marginal coverage of an interval, however, does not fully capture the quality of the predicted conditional quantiles. The typical argument is that we want an interval that has the correct marginal coverage while being as narrow, or sharp, as possible \citep{kuleshov2018Accurate}. A similar argument has been made in terms of calibration and refinement \citep{degroot1983Comparison} for a probabilistic classifier. This argument translates well to quantile regression.

 Suppose we are predicting a conditional $\tau$-quantile, denoted with $\hat{y}(X)$. The perfect predicted quantile would satisfy:
\[
\PR{Y < \hat{y}(X) \mid X = \bm{x}} = \tau \quad \forall \bm{x}.
\]
The predicted quantile is never perfect and we therefore make the following errors:
\begin{align}
\PR{Y < \hat{y}(X) \mid X = \bm{x}} = \tau &+ \left(\PR{Y < \hat{y}(X)} - \tau \right) \label{eq: calibration error} \\ 
 &+ \left(\PR{Y < \hat{y}(X) \mid X = \bm{x}} - \PR{Y < \hat{y}(X)}\right). \label{eq: refinement error}
\end{align}
The error in line \eqref{eq: calibration error}, $\PR{Y < \hat{y}(X)} - \tau$, is the calibration error. Crucially, this error can be low by having conditional quantiles that are only correct on average and not for individual values of $\bm{x}$. This can be seen by noting that:

\begin{equation*}
\PR{Y < \hat{y}(X)} = \int_{\mathcal{X}} \PR{Y < \hat{y}(X) \mid X=\bm{x}} \pi(\bm{x})d\bm{x},
\end{equation*}
where $\pi(\bm{x})$ is the density function of the random variable $X$.

The second error term, in line \eqref{eq: refinement error}, is the refinement error. This term is large if the coverage of the conditional quantile is substantially larger or smaller than the marginal coverage.

 As an example, the empirical CDF would be relatively well-calibrated but would not be practical as it entirely ignores the covariates. A similar point is made by \citet{kuleshov2018Accurate}. 
 
 2. The average pinball loss:
 
Because of the limitation of only reporting the marginal coverage, we also report the average pinball loss:
\[
\text{Average pinball loss} =	\frac{1}{N_{\tau} N} \sum_{i=1}^{N} \sum_{j=1}^{N_{\tau}} L_{\tau_{j}}(y_{i}, \hat{y}_{ij}),
\]
where $L_{\tau_{j}}$ is the pinball loss for quantile $\tau_{j}$, $N_{\tau}$ is the number of predicted quantiles, $N$ is the number of data points, and $\hat{y}_{ij}$ is the $j$-th predicted quantile of data point $i$.  The pinball loss is a proper scoring rule for conditional quantiles \citep{gneiting2007Strictlya} and therefore measures both the calibration error and the refinement error.

3. The crossing percentage, which is the percentage of adjacent predicted quantiles that cross:

\[
	\text{Crossing percentage}= \frac{1}{(N_{\tau}-1) N} \sum_{i=1}^{N} \sum_{j=1}^{N_{\tau}-1} \mathbb{I}_{\{\hat{y}_{ij} > \hat{y}_{i(j+1)} \}} \cdot 100 \%.
\]

\subsection{Results and Discussion}
\begin{figure}[h!]
\includegraphics[width=\linewidth]{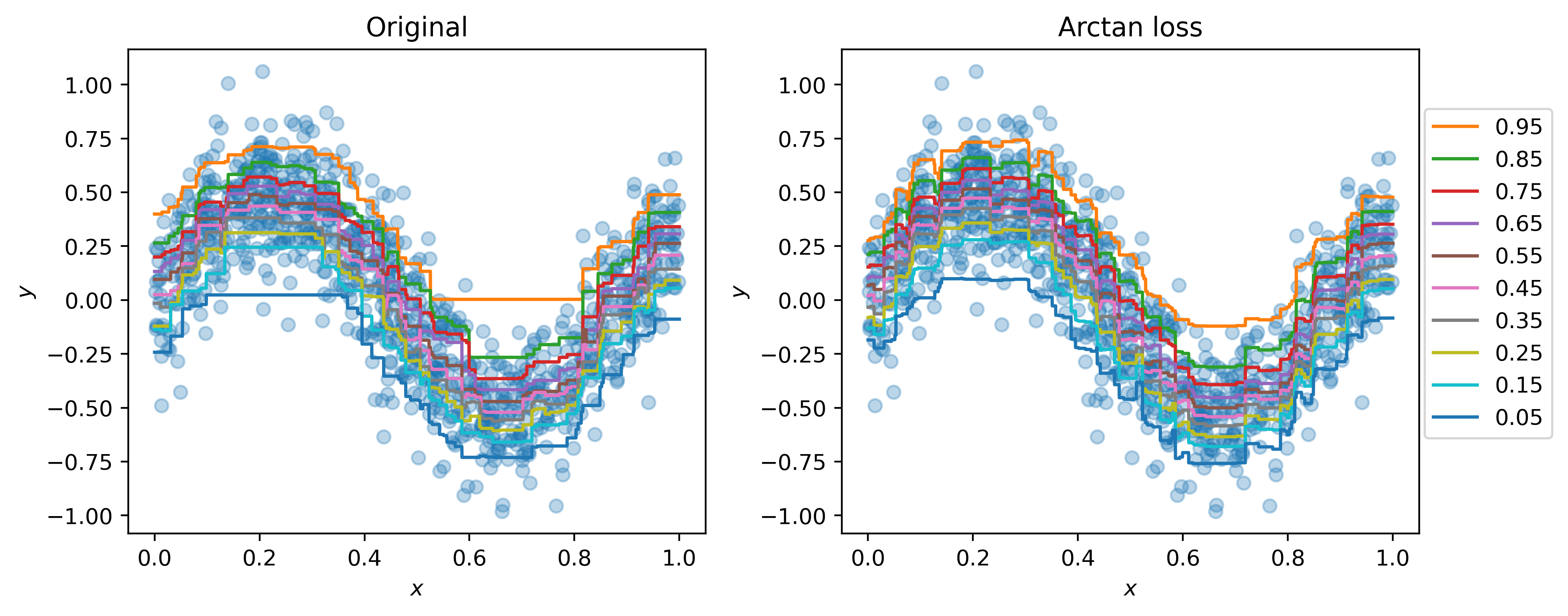}
\caption{By using the arctan loss function, a multi-output tree can be used in XGBoost. This causes the splits to be at the same locations, which reduces the number of crossings.}
\label{fig: sinexample}	
\end{figure}

\paragraph{Toy example} Figure \ref{fig: sinexample} illustrates the difference between our approach and the default implementation of quantile regression in XGBoost. The default implementation uses a separate model for each quantile. This causes the splits to be at different locations, easily resulting in quantile crossings. On the contrary, our approach uses a single model for all ten quantiles and therefore has the splits at the same locations. In this example, our approach had 0 crosses.

\paragraph{UCI data sets}
 The results on the six UCI data sets are given in Table \ref{table: UCI_qr_results}. We evaluated two values of $s$, our smoothing parameter. A larger value means more smoothing. 
 
 We observe far fewer crossings for all six data sets. Additionally, while our intervals are typically smaller, our marginal coverage is overall closer to the desired 90$\%$, with the exception of the energy data set. We do not see a clear difference in performance in terms of the pinball loss. This illustrates the previously mentioned fact that the coverage is a marginal coverage. A model can be very well calibrated, but not very informative, or it can be very informative yet poorly calibrated.
 
 The most dramatic example can be found in the wine data set. The 90$\%$ PI of the original approach only has a 61.2$\%$ marginal coverage. At first glance, this seems terrible. However, the average pinball loss shows that the actual model is not that much worse than our implementation. When investigating this further, we found that the original $0.05$-quantiles were slightly, but consistently, too large. This resulted in a very low coverage of the $90\%$ PIs even though the intervals were in fact only slightly too small. The problem is that they were consistently too small for multiple values of $\bm{x}$, resulting in an extremely low marginal coverage.

\begin{table}[tb]
\centering
\begin{adjustbox}{width=\linewidth}
\begin{tabular}{l|c c c| c c c| ccc| c cc}
\hline
\textbf{Data set} & \multicolumn{3}{|c|}{\textbf{Average Pinball loss}} & \multicolumn{3}{|c|}{$90 \%$ PI \textbf{coverage}} &  \multicolumn{3}{|c|}{\textbf{$90\%$ PI width}} &  \multicolumn{3}{|c}{\textbf{Crossing percentage}}\\ 
		&s=0.05 &s=0.1  & default & s=0.05 & s=0.1 & default & s=0.05 & s=0.1 & default& s=0.05 & s=0.1 & default \\\hline
Energy & 0.22  & 0.22 & 0.17 & 93.9 & 97.1 & 88.9& 5.4 & 5.8 & 4.5 & 7.1 & 0.3 & 20.2 \\
Concrete & 1.5     & 1.4  & 1.5  & 81.4 & 86.1 & 81.3  &15.2   &   15.8   &  19.8       & 5.5 & 3.2 &  16.1 \\
Kin8nm &0.040  & 0.040  & 0.041 &83.1 & 84.3 & 82.3 & 0.44 & 0.44 & 0.46 & 1.1 & 0.6 &  11.0 \\
Boston Housing & 0.92  & 0.91 & 0.95   & 80.0& 83.2 & 76.1  & 8.0   & 8.7  & 9.9  & 2.4 & 0.7 & 18.5  \\
Yacht & 0.21 & 0.26  & 0.21 & 90.9& 95.5& 69.2 & 2.8 & 4.9 & 5.5 & 0.5 & 0.0 & 29.8 \\
Wine & 0.17 & 0.17 & 0.17 & 88.1 & 88.7& 61.2& 1.5  & 1.7 & 1.8 & 3.1 & 2.2 & 5.3 \\
\end{tabular}
\end{adjustbox}
\caption{Results on UCI benchmark data sets. The marginal coverage and the average width are calculated for the 90$\%$ PIs that are constructed with the 0.05- and the 0.95 quantiles. Using the arctan pinball loss results in significantly fewer quantile crossings while achieving comparable performance. On 5 of the 6 data sets, energy being the exception, we observe superior coverage and an equal or better pinball loss. We also see the effect that using a larger value of $s$ results in more conservative quantiles and thus wider prediction intervals.}
\label{table: UCI_qr_results}
\end{table}
While the marginal coverage is often closer to 90$\%$ with the arctan loss, we also have a number of data sets where the intervals are too narrow, especially when using a smaller $s$. As mentioned, the original implementation even had a marginal coverage as low as 61.2$\%$ for one of the data set. This overconfidence is in line with the observation of \cite{guo2017calibration} who noted that modern machine learning models are often overconfident. The pinball loss depends on both calibration and refinement and therefore the resulting optimal model according to the pinball loss may not be the best calibrated model.

A general approach to improve the calibration is to add a post-hoc calibration step. The PIs are evaluated on a previously unseen part of the data set and are tuned such that they are better calibrated. We advise to always consider using such a post-hoc calibration step when implementing these models in practice. An example of such a procedure can be found in \cite{romanoConformalizedQuantileRegression2019}.

\paragraph{Electricity-grid substations}
The results for the electricity substations are given in Table \ref{table: cableresults}. Our approach yields comparable results while having far fewer crossing and requiring only a single model. For two of the four substations, we have a slightly better pinball loss while for two others, we perform slightly worse. A similar pattern is observed for the coverage.

\begin{table}[tb]
\centering
\begin{adjustbox}{width=\linewidth}
\begin{tabular}{l| c c|  c c| cc|  cc}
\hline
\textbf{Data set} & \multicolumn{2}{|c|}{\textbf{Average Pinball loss}} & \multicolumn{2}{|c|}{$90 \%$ PI \textbf{coverage}} &  \multicolumn{2}{|c|}{\textbf{$90\%$ PI width}} &  \multicolumn{2}{|c}{\textbf{Crossing percentage}}\\ 
		& $s=0.1$ & default &$s=0.1$ & default &$s=0.1$ & default& $s=0.1$ & default \\\hline
Substation 287 & 0.19 & 0.18 &80.6 & 76.7 & 1.66 & 1.56 & 0.4 & 12.8 \\
Substation 307 & 0.75 & 0.77 & 88.2 & 84.11 & 8.24 & 7.95 & 0.0 & 4.2 \\
Substation 435 & 0.53 & 0.55 & 78.8 & 86.1 & 4.83 & 5.88 & 0.0 & 2.1 \\
Substation 438 & 0.69 & 0.68 & 84.6 & 86.1 & 7.0 & 7.3 & 0.0 & 3.3 \\
\end{tabular}
\end{adjustbox}
\caption{Results on Alliander data sets. We obtain comparable performance with only a single model and have almost zero quantile crossings. The coverage is lower for both approaches, an effect we expect to be the result of the inability of tree based models to extrapolate well.}
\label{table: cableresults}
\end{table}

We also observed that the models are sometimes biased, rather than overconfident, causing a subpar coverage. We suspect that this is caused by the varying conditions combined with the fact that trees do not extrapolate well. 

For the first substation, the training data (including validation) starts halfway through October and ends in early January. The entire test set consists of days in January. During that time, the loads in the substation were typically lower. Multiple factors might cause this, but a likely explanation is reduced sunlight in January. 

\begin{figure}[h!]
\vskip -0.1in 
\includegraphics[width=0.49\textwidth]{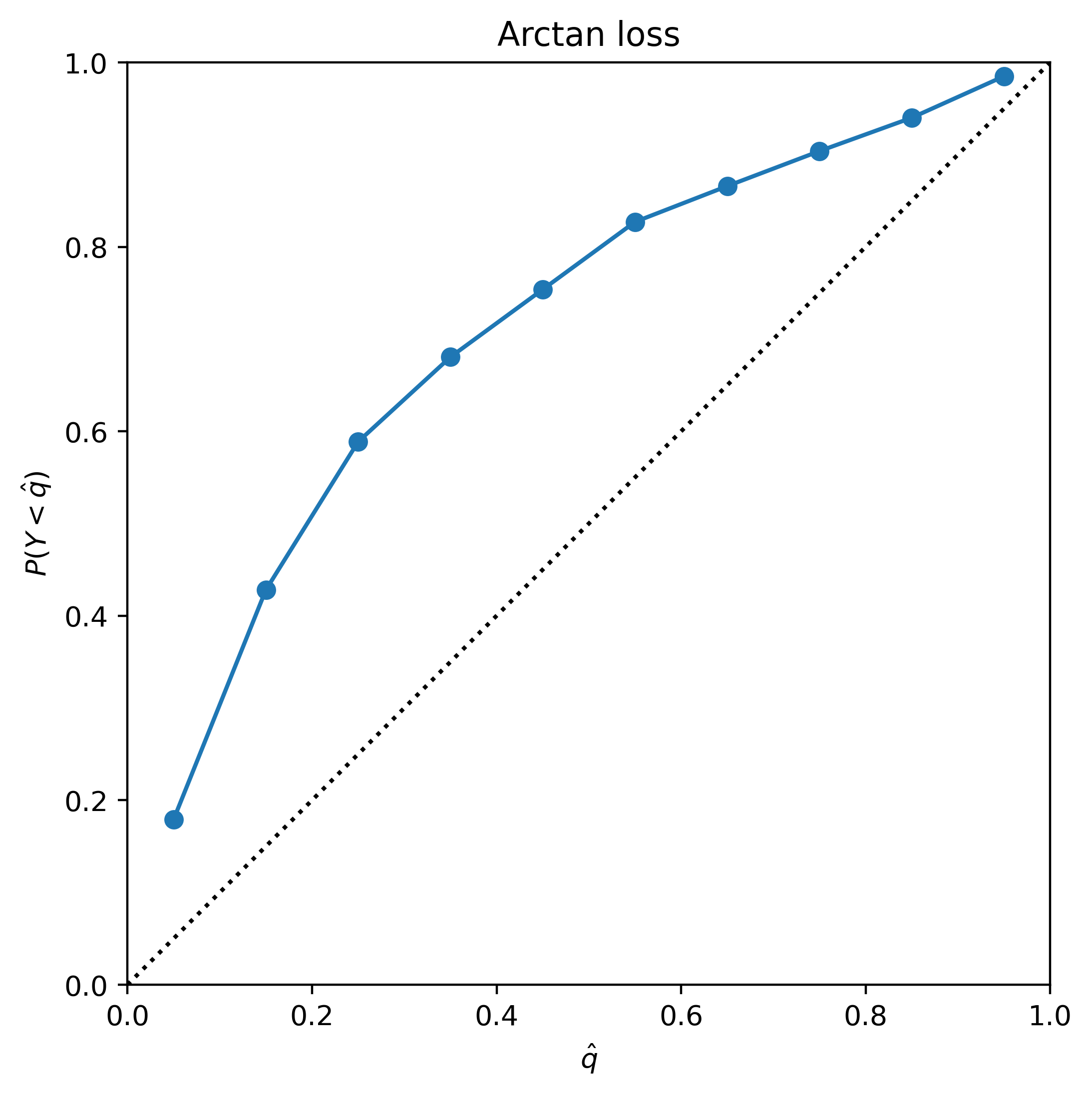}
\includegraphics[width=0.49\textwidth]{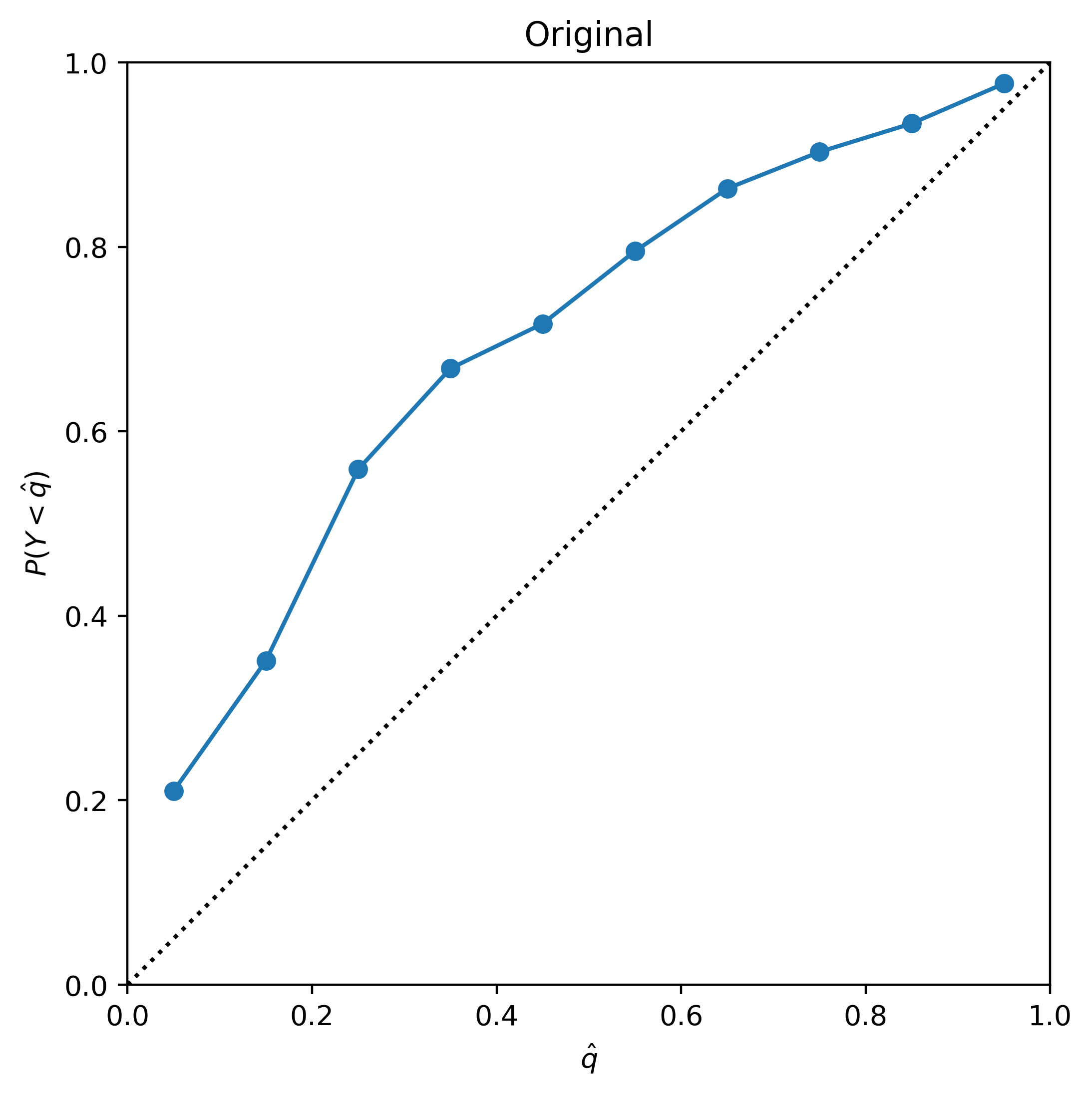}
\caption{Reliability diagrams for the quantiles obtained by using the arctan loss (left) or the regular pinball loss (right) on substation 287. Both models are biased, all the quantiles are too large, likely the result of a distributional shift between the training and test data.}
\label{fig: 287reliability}
\end{figure}

\begin{figure}[h!]
\vskip -0.2 in 
\includegraphics[width=0.49\textwidth]{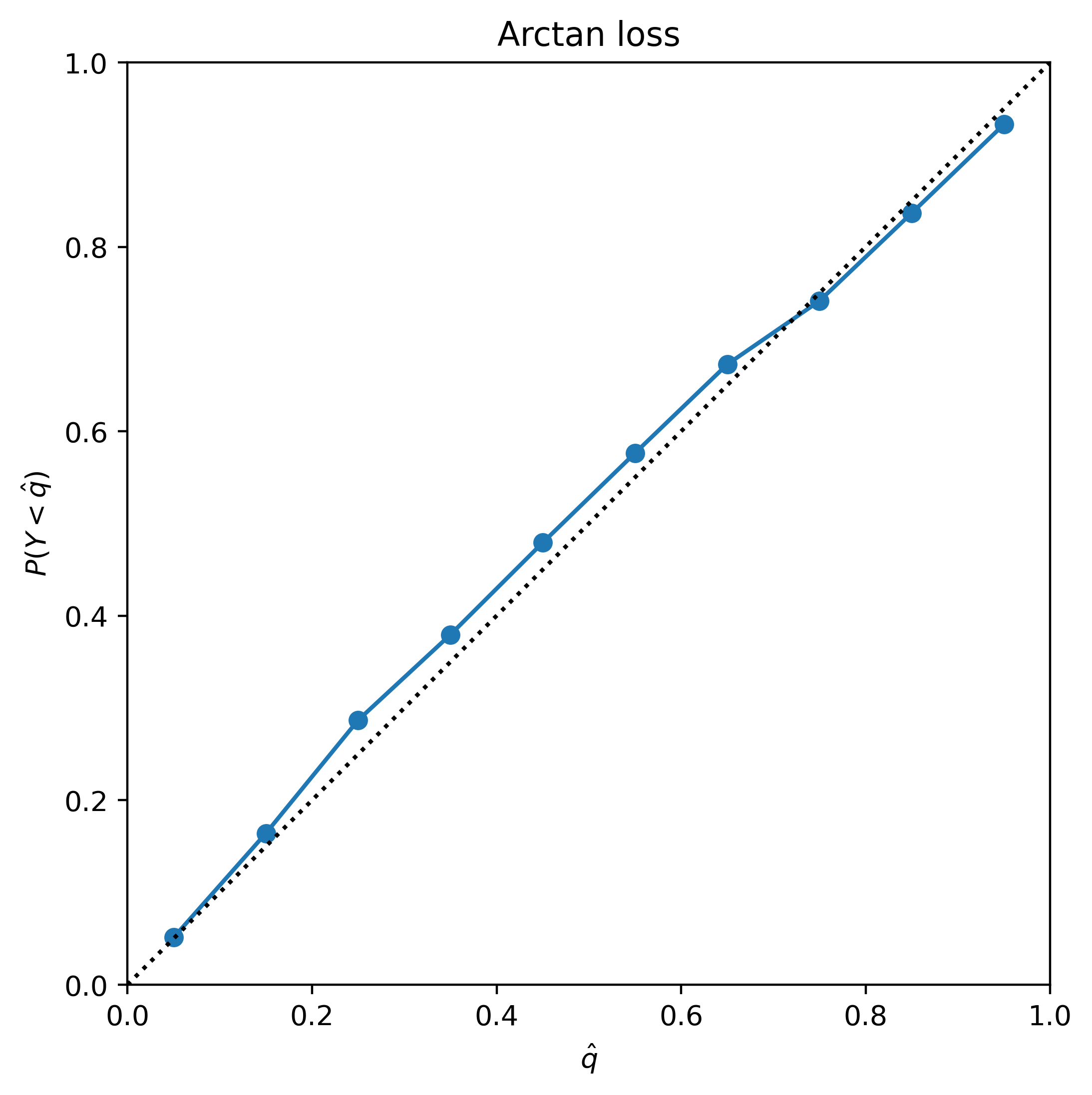}
\includegraphics[width=0.49\textwidth]{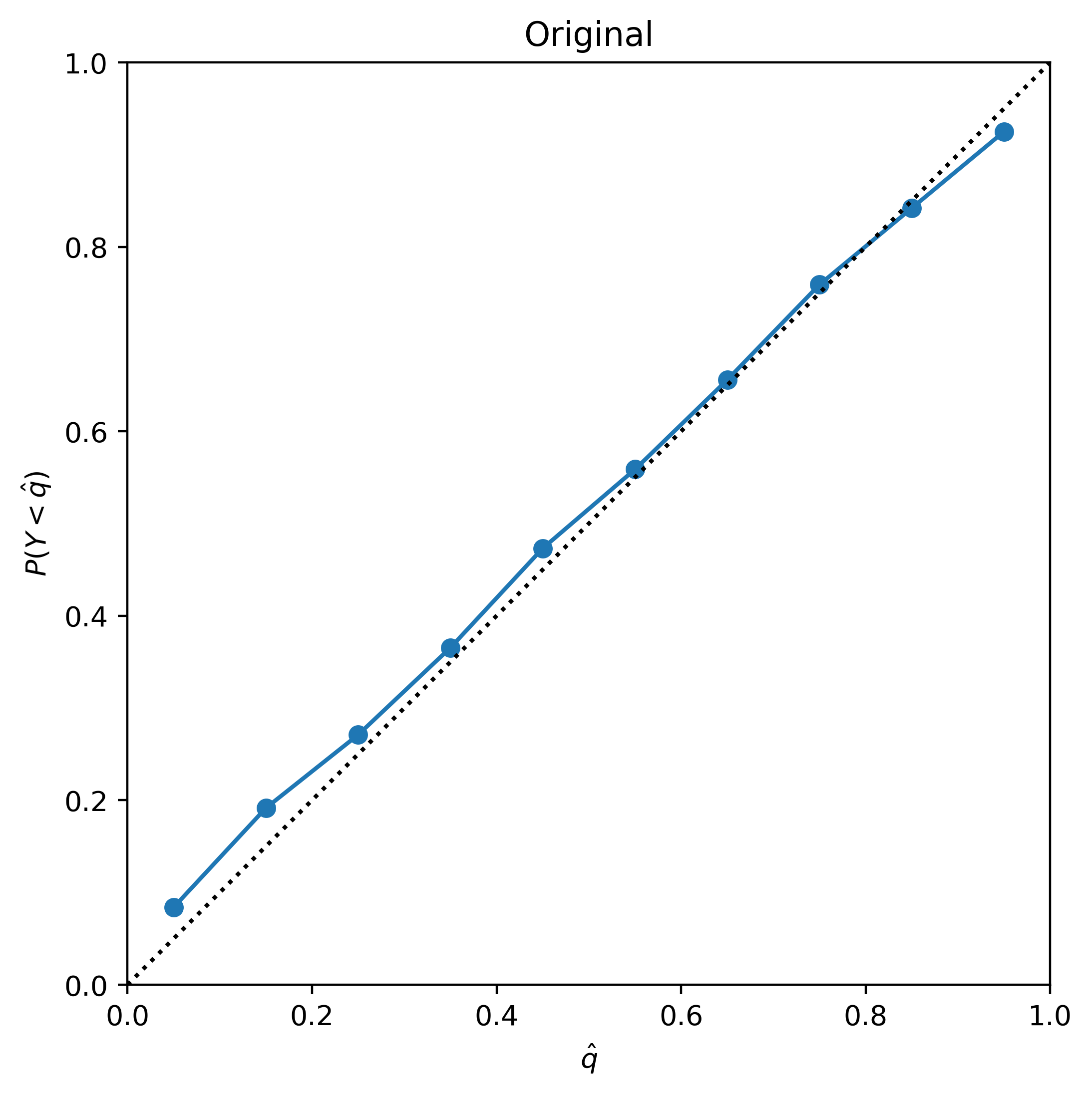}
\caption{Reliability diagrams for the quantiles obtained by using the arctan loss (left) or the regular pinball loss (right) on substation 307. Both models are relatively well-calibrated.}
\label{fig: 307reliability}
\end{figure}

The reliability diagrams in Figure \ref{fig: 287reliability} visualize this bias for substation 287. Reliability diagrams display the marginal performance of the conditional quantiles \citep{murphy1977Reliability, niculescu2005predicting}. All the quantiles were too large. For substation 307, visualized in Figure \ref{fig: 307reliability}, we observe that both approaches yielded well-calibrated quantiles.

\begin{figure}[h]
\includegraphics[width=0.99\textwidth]{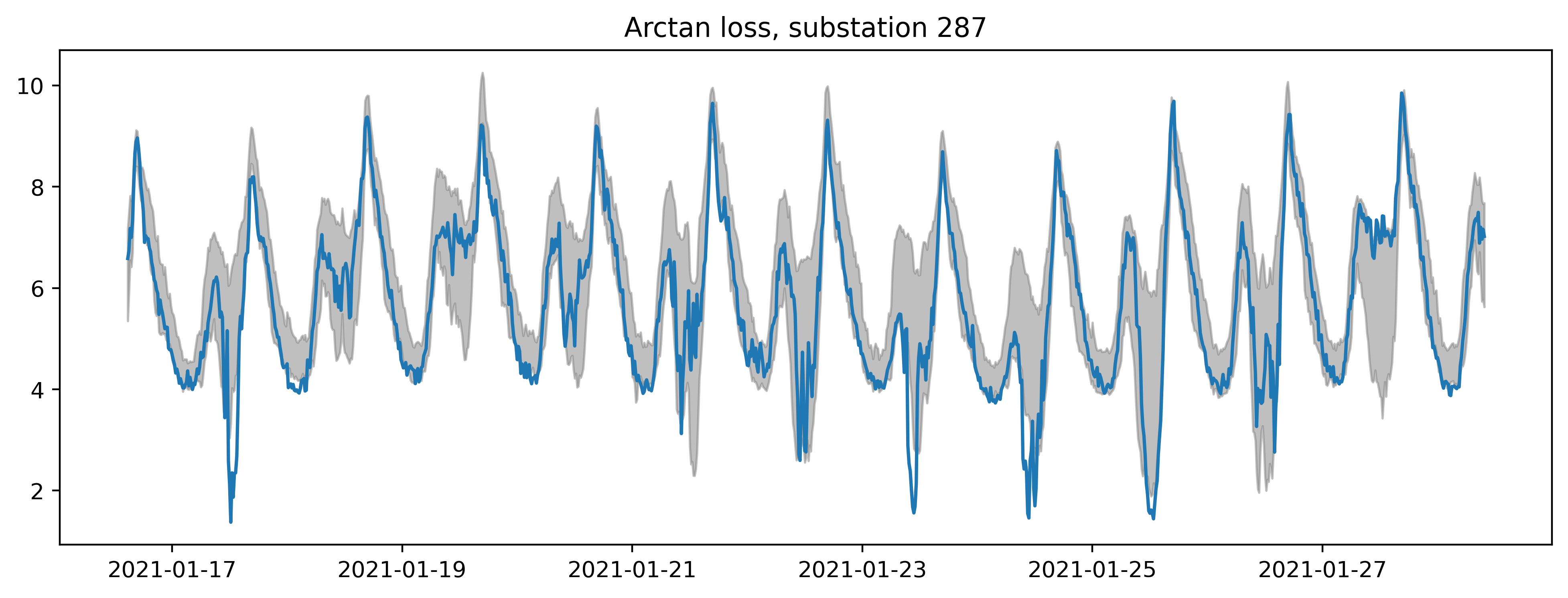}
\includegraphics[width=0.99\textwidth]{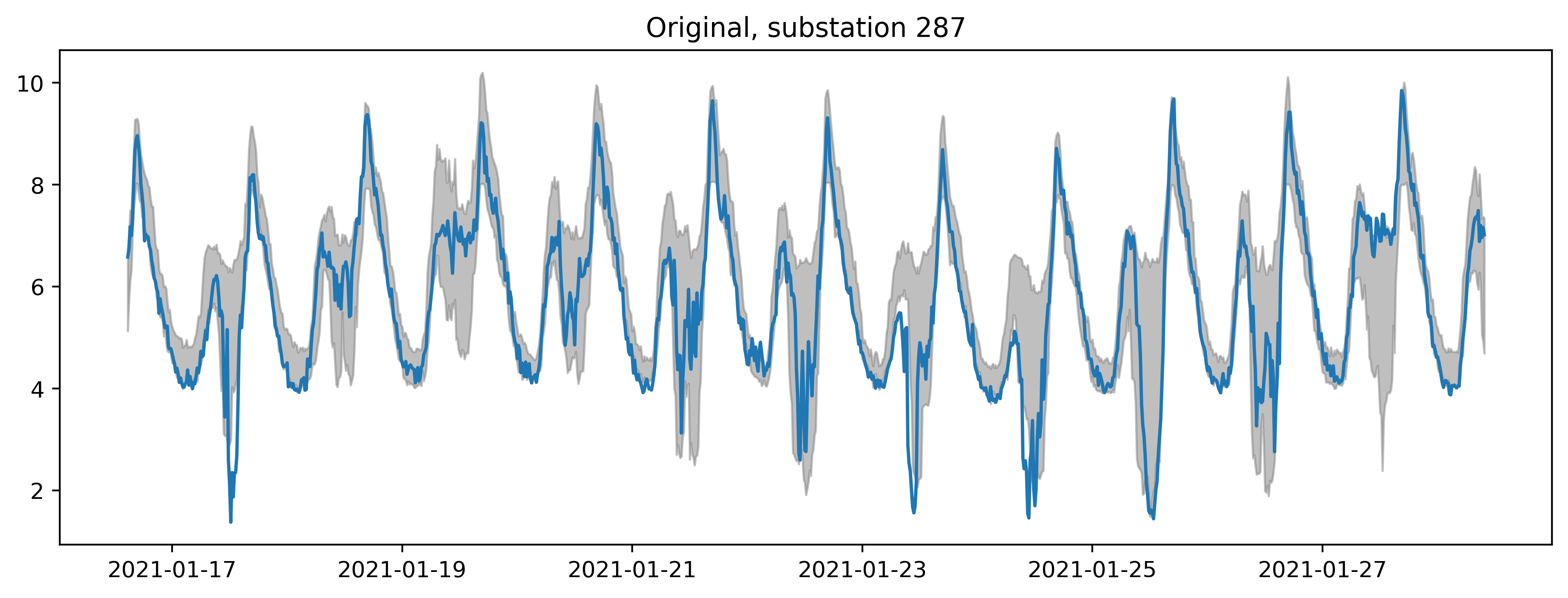}
\includegraphics[width=0.99\textwidth]{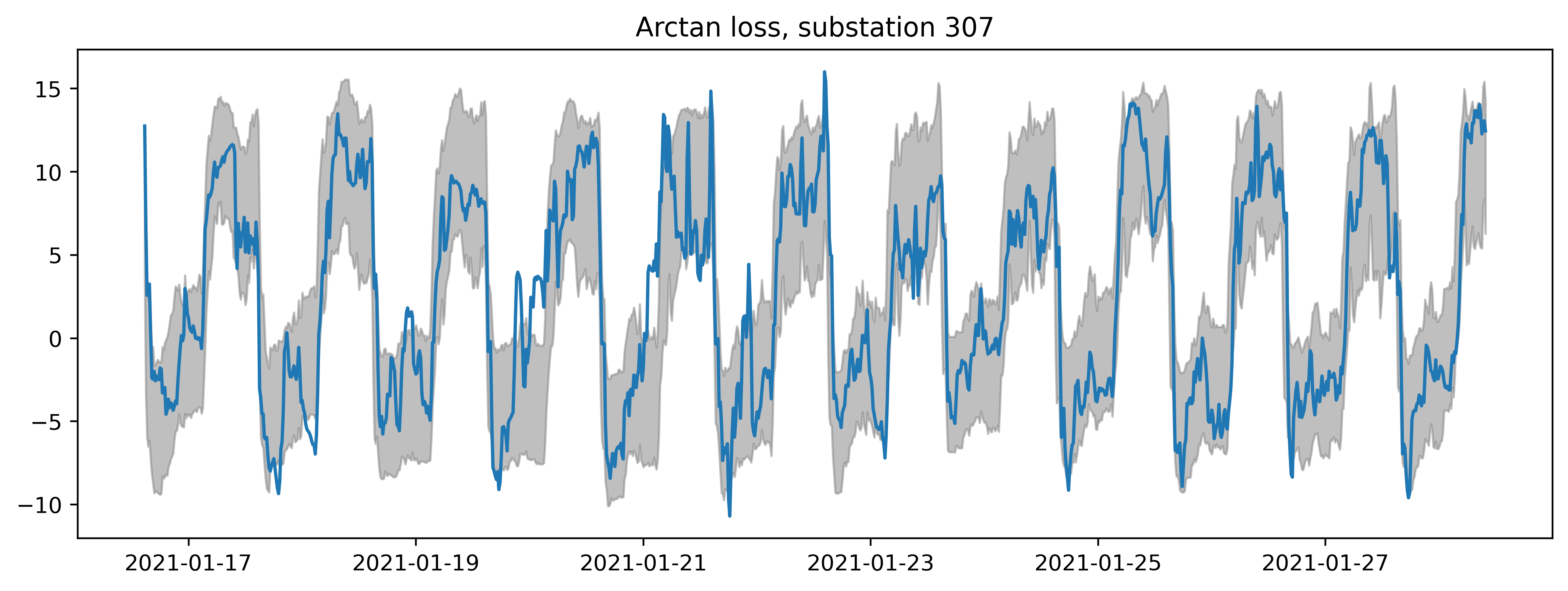}
\includegraphics[width=0.99\textwidth]{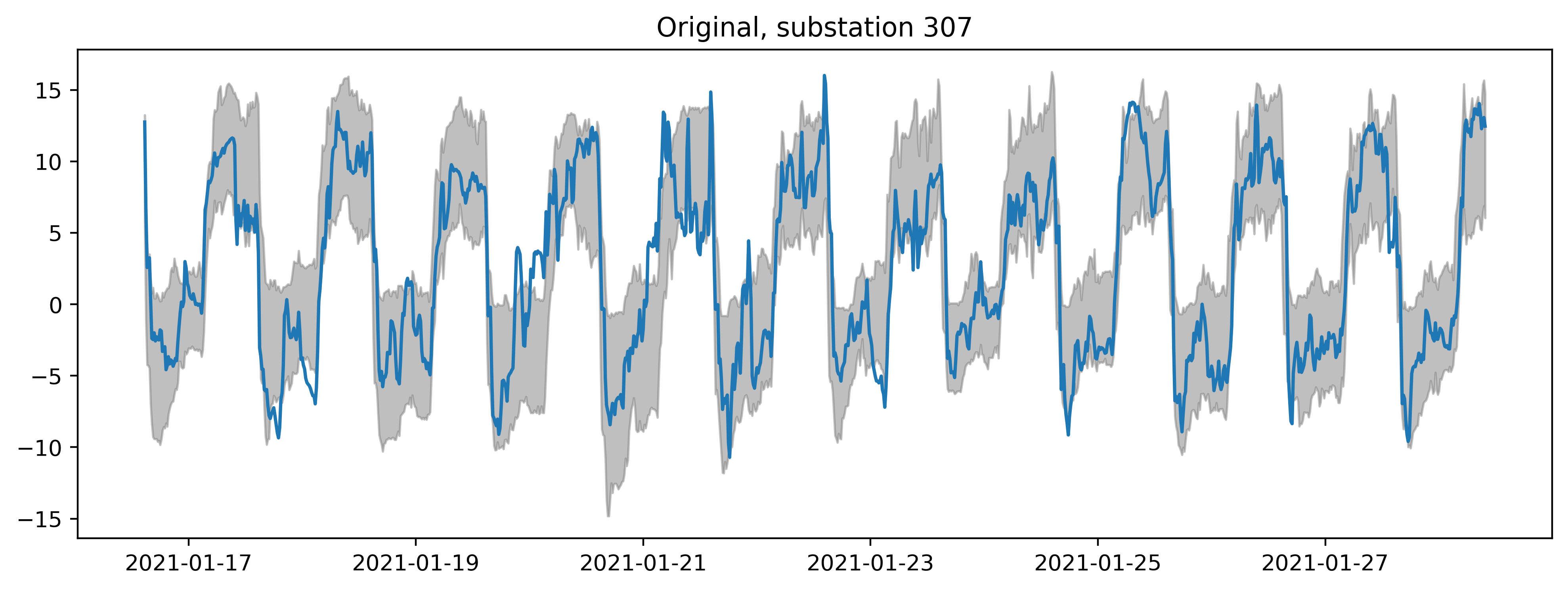}
\caption{A visualisation of the time series for substations 287 and 307. The grey area gives the 90$\%$ PI that is constructed using the 0.05- and 0.95-quantiles. For substation 287, the values of the lowest peaks are overestimated by both models and there is a substantial bias around the 24th of January.}
\label{fig: timeseries}
\end{figure}

The $90\%$ PIs for substations 287 and 307 are visualised for both approaches in Figure \ref{fig: timeseries}. For substation 287, we observe that the intervals fail to capture the lowest peaks. Additionally, both models exhibit a bias around the 24th of January. The predicted intervals are often above the actual loads in that region. 

\FloatBarrier

In this specific application, which involves a time series with only three months of training data, using tree-based models presents clear disadvantages. Inevitably, the model will encounter unseen scenarios, such as the first significantly sunny day or the first frost period in a three-month period. In these instances, the model may fail since the individual trees cannot extrapolate.

This illustrates that XGBoost may not always be the correct model. We stress that we do not claim this to be the optimal way to perform quantile regression. Other models, such as neural networks or even simple linear models, could work just as well, or better, depending on the specific situation. However, there may be situations where XGBoost is preferred due to its efficient training and its capability to handle missing data. In such cases, using the arctan pinball loss allows for the simultaneous estimation of multiple quantiles, resulting in substantially fewer crossings.


\section{Conclusion} \label{conclusionsection}
This study introduced a novel smooth approximation of the pinball loss function, termed the arctan pinball loss, which has been specifically designed to meet the needs of the XGBoost framework. The key advantage of this loss function lies in its second derivative, which decreases significantly more slowly than that of the currently available alternatives. 

This arctan loss facilitates the use of a single model for multiple quantiles simultaneously. This is both more efficient and greatly reduces the number of quantile crossings. The experimental results demonstrate that this approach is viable for a wide range of data sets and yields competitive results while using only a single model and while having far fewer quantile crossings.

%

\FloatBarrier
\bibliographystyle{apalike}
\bibliography{../../references4}
\newpage
\appendix
\begin{center}
\vskip 0.2in
\large{APPENDIX}	
\end{center}

\section{Constructing the arctan pinball loss} \label{arctanconstruction}
Recall that we defined $u:= y - \hat{y}$. We will discuss the pinball loss as a function of $u$ and $\tau$, the desired quantile. The classical pinball loss is $(\tau-1)u$ for $u<0$ and $\tau u$ for $u>0$. We can place this pinball loss in a larger set of functions, namely $\{L(u)=(\tau - f(u))u\}$. For the pinball loss, $f(u)$ is a stepfunction that goes from 1 to 0 at $u=0$.

We can also consider other functions for $f(u)$ that go from 1 to 0 to find approximations of the loss function. For a loss function to be suitable for XGBoost we require that $L''(u)$ is non-negligible in the relevant domain of $u$. If the targets are standardized, this relevant domain is roughly from -10 to 10 for most data sets.

It is immediately clear that the classical pinball loss is not suitable. 
A simple calculation shows that 
\[
L''(u) = -2 f'(u) - f''(u)u.
\]
Writing it out in terms of $f(u)$ allows us to easily check different functions and see how quickly the second derivative goes to zero.

We propose the following function $f$ as a suitable candidate: 
\[
	f_{s}(u) = 0.5 - \frac{\arctan(u/s)}{\pi}.
\]
Using this  $f$ would result in the following loss function:
\[
L_{\tau, s}(u) = (\tau - 0.5 + \frac{\arctan (u/s)}{\pi})u.
\]
However, this loss function is asymptotically biased, as we demonstrate for the limit $u \rightarrow \infty$. The limit $u$ to $-\infty$ is identical and can be obtained similarly.
\begin{align*}
\lim_{u \rightarrow \infty}L_{\tau, s}(u) - L^{\text{pinball}}_{\tau}(u)   &= \lim_{u \rightarrow \infty}  (\tau - 0.5 + \frac{\arctan (u/s)}{\pi})u - \tau \cdot u \\
&=\lim_{u \rightarrow \infty} (-0.5 +  \frac{\arctan (u/s)}{\pi})u \\
&= \lim_{u \rightarrow \infty} \frac{(-0.5 +  \frac{\arctan (u/s)}{\pi})}{u^{-1}} \\
&\overset{\text{L'H\^opital}}{=} \lim_{u \rightarrow \infty} -\frac{\frac{1}{\pi s} \frac{1}{1+(u/s)^{2}}}{u^{-2}} \\
&= -\frac{1}{\pi s}  \lim_{u \rightarrow \infty} \frac{u^{2}}{1 + (u/s)^{2}} \\
&= -\frac{1}{\pi s}  \lim_{u \rightarrow \infty} \frac{s^{2}u^{2}}{s^{2} + u^{2}} \\
&= -\frac{s}{\pi}.
\end{align*}
To obtain an asymptotically unbiased loss function, we therefore add a $\frac{s}{\pi}$ term and end up with our arctan pinball loss:
\[
L_{\tau, s}^{\text{arctan}}(u) = (\tau - 0.5 + \frac{\arctan (u/s)}{\pi})u + \frac{s}{\pi}.
\]
Crucially, the second derivative of this arctan pinball loss is polynomial:
\begin{align*}
\frac{\partial^{2} L^{\text{arctan}}_{\tau, s}(u)}{\partial u^{2}} &= 	\frac{2}{\pi s}\left(1 + (u/s)^{2}\right)^{-1} - \frac{2u^{2}}{\pi  s^3}  \left(1 + (u/s)^{2}\right)^{-2} \\
&= \frac{2}{\pi s}(1 + (u/s)^{2})^{-2}.
\end{align*}

\end{document}